\documentclass{article}

\usepackage{arxiv}

\usepackage{lineno,hyperref}
\usepackage{epsfig}
\usepackage{graphicx}
\usepackage{amsmath}
\usepackage{amssymb}
\usepackage{multirow}
\usepackage{caption}
\usepackage[ruled,vlined]{algorithm2e}
\usepackage{tabularx}
\newcolumntype{Y}{>{\centering\arraybackslash}X}
\usepackage{makecell}
\modulolinenumbers[5]
\usepackage[utf8]{inputenc} 
\usepackage[T1]{fontenc}    
\usepackage{hyperref}       
\usepackage{url}            
\usepackage{booktabs}       
\usepackage{amsfonts}       
\usepackage{nicefrac}       
\usepackage{microtype}      
\usepackage{lipsum}		
\usepackage{graphicx}
\usepackage{natbib}
\usepackage{doi}

\title{Entropy-Aware Model Initialization for \\Effective Exploration in Deep Reinforcement Learning}

\author{
    Sooyoung Jang \\
	Electronics and Telecommunications Research Institute (ETRI)\\
	Daejeon, South Korea    \\
	\texttt{sy.jang@etri.re.kr} \\
	\And
	Hyung-Il Kim\thanks{Correspongding author} \\
	Electronics and Telecommunications Research Institute (ETRI)\\
	Daejeon, South Korea    \\
	\texttt{hikim@etri.re.kr} \\
}

\renewcommand{\shorttitle}{\textit{arXiv} Template}

\hypersetup{
pdftitle={Entropy-Aware Model Initialization for Effective Exploration in Deep Reinforcement Learning},
pdfsubject={q-bio.NC, q-bio.QM},
pdfauthor={S.Jang, H.Kim},
pdfkeywords={Deep reinforcement learning, entropy, model initialization, exploration},
}

\begin{document}
\maketitle

\begin{abstract}
Encouraging exploration is a critical issue in deep reinforcement learning. We investigate the effect of initial entropy that significantly influences the exploration, especially at the earlier stage. Our main observations are as follows: 1) low initial entropy increases the probability of learning failure, and 2) this initial entropy is biased towards a low value that inhibits exploration. Inspired by the investigations, we devise entropy-aware model initialization, a simple yet powerful learning strategy for effective exploration. We show that the devised learning strategy significantly reduces learning failures and enhances performance, stability, and learning speed through experiments.
\end{abstract}

\keywords{Deep reinforcement learning \and entropy \and model initialization \and exploration}

\section{Introduction}
Reinforcement learning has a long history as an optimization technique for solving sequential decision-making problems \citep{gullapalli1990stochastic}. 
With the adoption of deep learning technology to reinforcement learning (so-called deep reinforcement learning (DRL)), it has shown successful performance even with high dimensional observation and action spaces in the fields of robotic control \citep{yang2018hierarchical,haarnoja2018composable}, game \citep{silver2017mastering, patel2019improved}, medical \citep{ghesu2017multi, raghu2017continuous} and financial \citep{zarkias2019deep,tsantekidis2021diversity} applications. 
In such a DRL framework, the exploration-exploitation trade-off is one of the crucial issues affecting the performance of the DRL algorithm \citep{ishii2002control}.
In other words, even if the exploitation that makes the best decision over the current information is successful, the solution obtained by the DRL would not be optimal without a number of explorations. 
We can encourage the exploration by considering the entropy, which is the key metric for measuring the stochasticity of the action selection. 
That is why many recent studies \citep{schulman2017proximal, haarnoja2018soft, seo2021state, zhang2018efficient,ahmed2019understanding} refer to entropy. 

In \citep{schulman2017proximal}, a proximal policy optimization algorithm was proposed, where the entropy bonus term was augmented to ensure sufficient exploration motivated by \citep{williams1992simple, mnih2016asynchronous}.
And, a soft actor-critic DRL algorithm based on the maximum entropy RL framework was proposed in \citep{haarnoja2018soft}, where the entropy term was incorporated to improve exploration by acquiring diverse behaviors in the objective with the expected reward. 
However, these approaches, which consider the entropy along with other factors (\textit{e.g.}, reward) in the objective, are hard to handle the low entropy at the model initialization.
In \citep{ahmed2019understanding}, the impact of entropy on policy optimization was extensively studied. The authors observed that a more stochastic policy (\textit{i.e.}, the policy with high entropy) improved the performance of the DRL. 
The authors in \citep{andrychowicz2021matters} analyzed the effect of experimental factors in the DRL framework, where the offset in the standard deviation of actions was reported as one of the important factors affecting the performance of DRL.
These studies deal with the continuous control tasks, where the initial entropy can be easily controlled by adjusting the standard deviation. 
As far as we know, for discrete control tasks, neither a research result reporting the effect of initial entropy nor a learning strategy exploiting it is not present.
One of the reasons may be the difficulty of controlling the entropy of the discrete control tasks.
The entropy in the discrete control task is determined by the action selection probability obtained through the rollout procedure, in contrast to the continuous control task, where the standard deviation determines the entropy.

In order to address the concerns above-mentioned, we conduct an experimental study to investigate the effect of the initial entropy focusing on the tasks with discrete action space. 
Based on the experimental observations, we devise a learning strategy for the DRL, namely Entropy-aware model initialization. 
Our contributions of the paper can be summarized as follows: 
\begin{itemize}
\item We reveal the root cause of the frequent learning failures despite the easy tasks. We investigate that the model with low initial entropy significantly increases the probability of learning failures but the initial entropy is biased towards a low value.
Moreover, we observe that the initial entropy varies depending on the task and model initialization. These dependencies make it hard to control the initial entropy of the discrete control tasks.

\item We devise Entropy-aware model initialization, a simple yet powerful learning strategy that exploits the effect of initial entropy that we have analyzed. 
The devised learning strategy repeats model initialization and entropy measurement until the initial entropy exceeds the entropy threshold.
It can be used with any reinforcement learning algorithm since it just provides the algorithm with a well-initialized model.
Experimental results show that the entropy-aware model initialization significantly reduces learning failures and improves performance and learning speed. 
\end{itemize}
In Section~\ref{sec:effect}, we show the results of the experimental study, which investigates the effect of initial entropy on the DRL performance with discrete control tasks. Then, in Section~\ref{sec:devised}, we describe the devised learning strategy and discuss the experimental results in Section~\ref{sec:exp}. Finally, we conclude our paper in Section~\ref{sec:con}.

\section{Effect of Initial Entropy in DRL}\label{sec:effect}
In order to investigate the effect of the initial entropy in the DRL framework, we adopted the policy gradient method (\textit{i.e.}, proximal policy optimization \citep{schulman2017proximal}) implementation in the RLlib \citep{liang2017ray}.
The network architecture was set to the same as in \citep{schulman2017proximal}.
For this experimental study, we consider the tasks with discrete action space from the OpenAI Gym \citep{brockman2016openai}.

\begin{figure}[!t]
  \centering
 \includegraphics[width=0.7\linewidth]{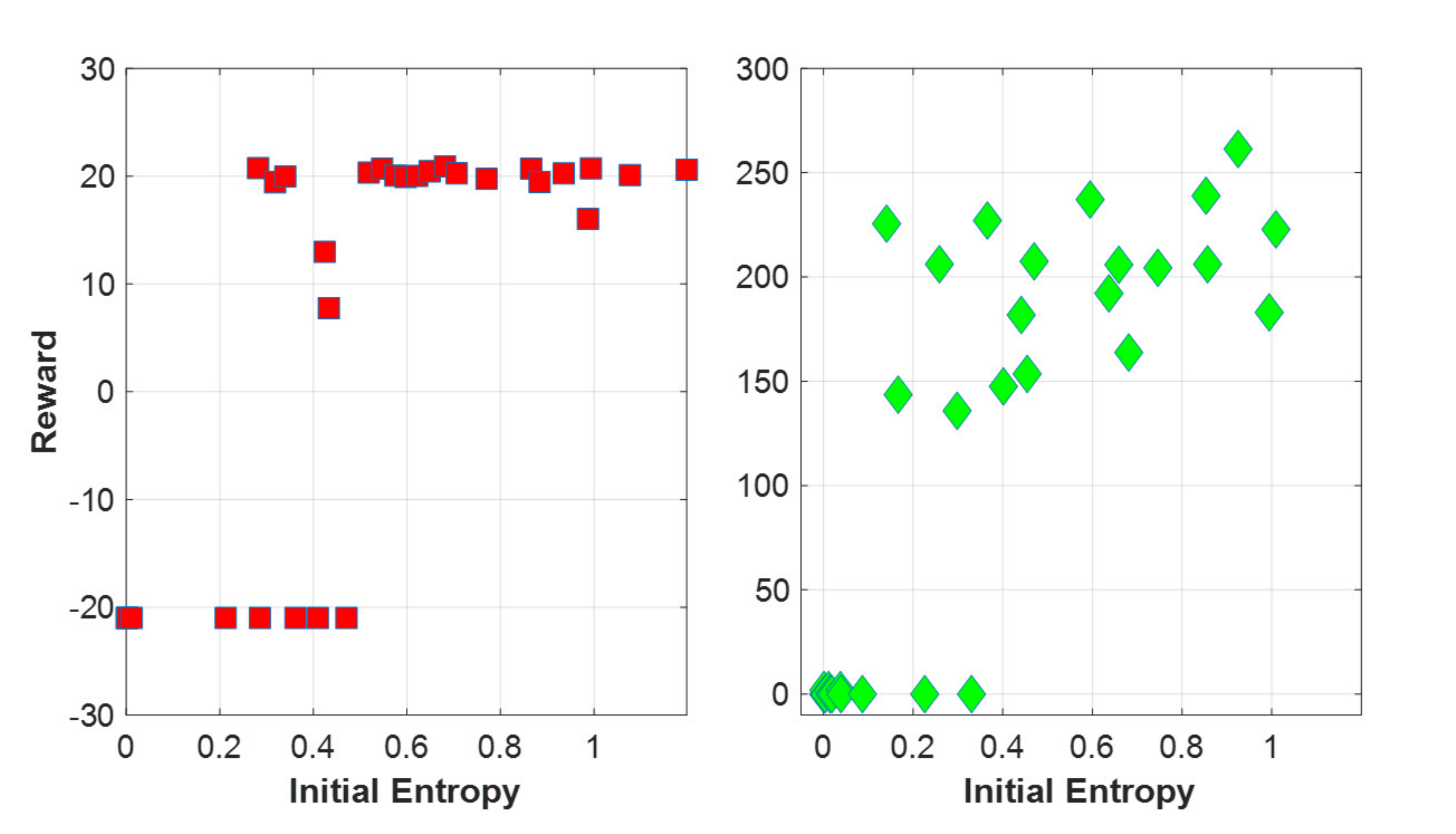}
 \caption{The reward depending on the initial entropy (left: Pong, right: Breakout), where 30 models were generated to investigate the effect of the initial entropy on the performance. }
\label{fig_reward}
\end{figure}

First, we investigate the effect of the initial entropy on the performance (\textit{i.e.}, reward). 
We generated 30 differently initialized models for the experiment and measured the rewards after 3,000 and 5,000 training iterations for Pong and Breakout, respectively.
For each iteration, 2,048 experiences were collected with 16 workers and 6 SGD epochs were made with a learning rate of 2.5e-4. 
Fig.~\ref{fig_reward} shows the reward for the initial entropy. 
We can see that the lower the initial entropy, the higher the learning failures (\textit{i.e.}, around -21 and 0 for Pong and Breakout, respectively). 
The low initial entropy leads to learning failures by inhibiting the exploration since it causes the agent to perform the same action for each state with a high probability.
It reminds us of the importance of exploration, especially at the earlier training stage.

\begin{figure}[!t]
  \centering
 \includegraphics[width=0.7\linewidth]{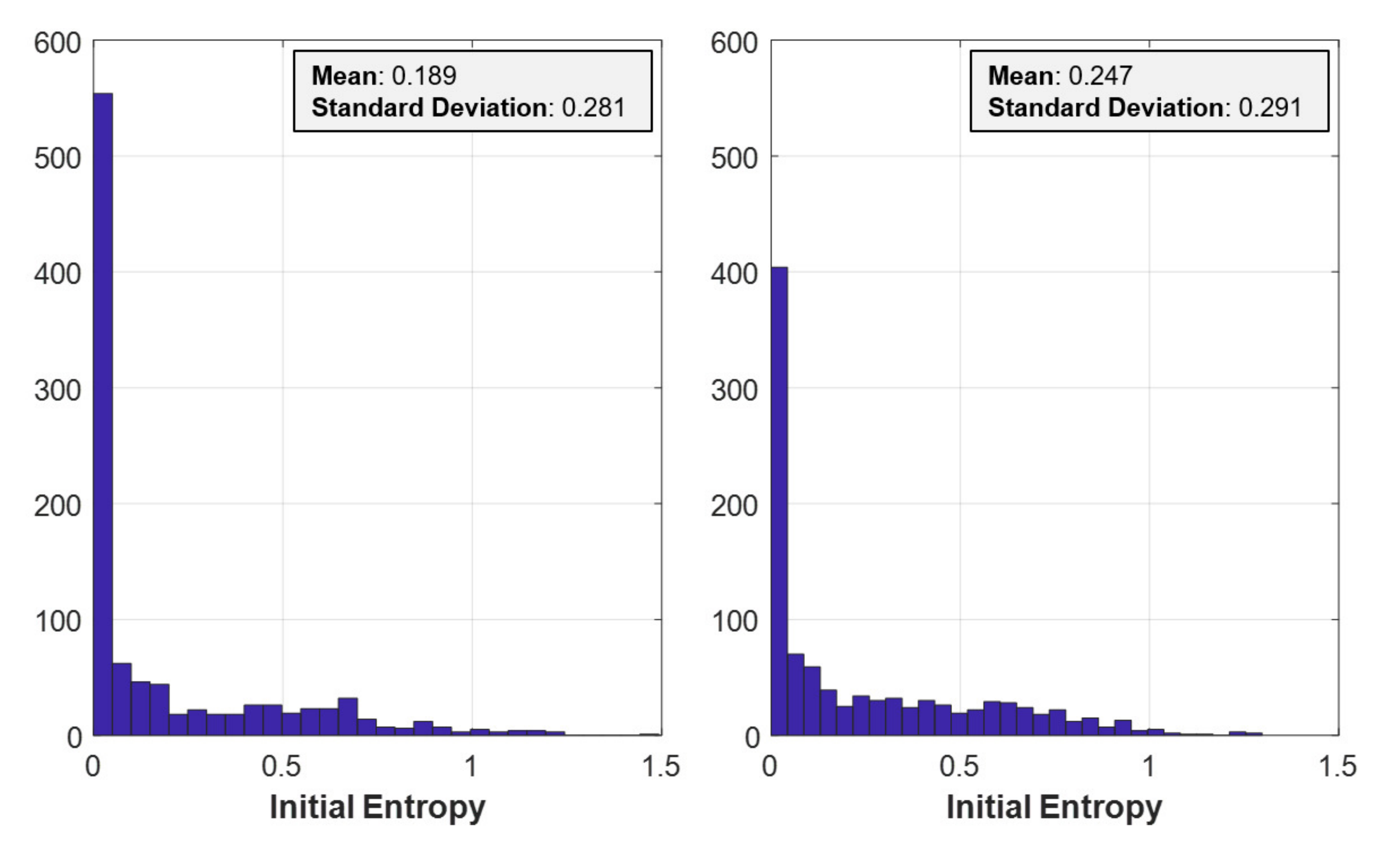}
 \caption{The histograms of the initial entropy with 1,000 models generated with different random seeds for two tasks (left: Pong, right: Breakout).}
\label{fig_dist}
\end{figure}

Then, we investigate the distribution of the initial entropy. For the investigation, we generated 1,000 models with different random seeds each for two tasks (\textit{i.e.}, Pong and Breakout) and measured the initial entropy values. 
Fig.~\ref{fig_dist} shows that the initial entropy is biased towards low values. 
Here, the average initial entropy values are 0.189 and 0.247 for Pong and Breakout, respectively. 
Our experimental findings (i.e., the high probability of learning failures for low initial entropy, and the low biased initial entropy) explain why the DRL often fails for tasks with discrete action spaces (Pong and Breakout as a case study) and why the performance drastically varies for each experiment.

Finally, we investigate the factors that affect the initial entropy. Table~\ref{tab:t1a} shows that both tasks and initial weights significantly impact the initial entropy.
We can see that the initial entropy varies with the task even with the same initial weight (\textit{e.g.}, Seed 05’s Atlantis and Breakout cases). 
In addition, the initial entropy differs by the initial weight of the model even with the same task (\textit{e.g.}, Seed 05 and Seed 09 in Atlantis). 
This is because the input image, which is the observation, greatly differs for each task.
These task and model initialization dependencies on the initial entropy make it hard to control the initial entropy.
\begin{table*}[!t]
\begin{center}
\caption{Initial entropy depending on the task with different size of action space under different random seeds, where ``STD'' denotes the standard deviation of the initial entropy values for 10 different random seeds.}
\label{tab:t1a}
\centering
\vspace{0.3cm}
\begin{tabular}{|c||c|c|c|c|}
\hline
 & \multicolumn{4}{c|}{Size of Action Space}\\ \cline{2-5}
 & \multicolumn{2}{c|}{4} & \multicolumn{2}{c|}{6} \\ \hline
\textbf{Task}                 & \textbf{Atlantis}   & \textbf{Breakout}  & \textbf{Pong}       & \textbf{Qbert}         \\ \Xhline{3\arrayrulewidth}
Seed 01       & 2.48e-01    & 2.95e-01   & 1.48e-03    & 4.74e-01         \\ 
Seed 02       & 8.63e-05    & 5.37e-12   & 9.68e-04    & 8.61e-01      \\ 
Seed 03       & 1.61e-02    & 3.38e-01   & 9.76e-01    & 2.70e-01      \\ 
Seed 04       & 1.32e-01    & 8.27e-05   & 7.20e-04    & 2.23e-02       \\ 
Seed 05       & 5.07e-01    & 6.65e-10   & 8.04e-01    & 1.58e-01       \\ 
Seed 06       & 2.30e-02    & 9.47e-02   & 8.98e-05    & 4.68e-01       \\ 
Seed 07       & 9.74e-09    & 7.13e-02   & 5.64e-01    & 1.58e-01       \\ 
Seed 08       & 2.55e-01    & 7.73e-01   & 1.18e-01    & 3.42e-01       \\ 
Seed 09       & 0.00e+00    & 2.18e-02   & 5.73e-01    & 4.34e-01       \\ 
Seed 10       & 6.62e-05    & 5.79e-01   & 1.73e-03    & 2.79e-01       \\ \Xhline{3\arrayrulewidth}
STD   & 0.171    & 0.274   & 0.385    & 0.233    \\ \hline
\end{tabular}
\end{center}
\end{table*}

From the above observations, the DRL algorithms are required to be learned by models with high initial entropy, and we need a strategy to generate such one.

\section{Entropy-Aware Model Initialization}\label{sec:devised}
According to Section~\ref{sec:effect}, we observe that 1) the learning failure frequently occurs with the model with low initial entropy, 2) the initial entropy is biased towards a low value, and 3) even with the same network architecture, the initial entropy greatly varies on the task and the initial weights of the models. 
Inspired by the experimental observations, we propose the entropy-aware model initialization strategy.
The learning strategy repeatedly initializes the model until the initial entropy value of the model exceeds the entropy threshold.
In other words, the proposed learning strategy is to encourage the DRL algorithms such as PPO \citep{schulman2017proximal} or SAC \citep{haarnoja2018soft} to collect a variety of experiences at the initial stage by providing the model with high initial entropy.
As can be seen in Algorithm~\ref{alg1}, suppose that the task ($E$), number of actors ($M$), entropy threshold ($h_{th}$), and initialized model ($\pi_{0}$) are given.
For each $j$-th actor, we first store action selection probabilities ($p_{\pi_{i}}^{(j,t)}$) from the rollout with the model ($\pi_{i}$) for each timestep $t\in\{1, \cdots, T\}$, where $T$ denotes the horizon. 
Note that the action selection probability for the set of action space $\mathcal{A}$ is the softmax of the outputs of $\pi_{i}$.
Next, the mean entropy ($\hat{h}_{\pi_{i}}$) is computed over the total action selection probabilities collected from the $M$ actors.
\begin{equation}\label{eqn1}
    \hat{h}_{\pi_{i}}=\frac{1}{MT}\sum_{j=1}^{M}\sum_{t=1}^{T}h_{\pi_{i}}^{(j,t)},
\end{equation}
\noindent where $h_{\pi_{i}}^{(j, t)}$ is defined by: 
\begin{equation}
    h_{\pi_{i}}^{(j, t)}=-\frac{1}{|\mathcal{A}|}\sum_{x\in \mathcal{A}}p_{\pi_{i}}^{(j,t)}(x)\,\mathrm{log}\,p_{\pi_{i}}^{(j,t)}(x).
\end{equation}
Then, we compare the $\hat{h}_{\pi_{i}}$ to the entropy threshold ($h_{th}$). If the $\hat{h}_{\pi_{i}}$ exceeds the $h_{th}$, then we stop and continue with the DRL algorithm.
If not, we set the random seed to a different value, reinitialize the model ($\pi_{i+1}$), and repeat the above process until the $\hat{h}_{\pi_{i}}$ exceeds the $h_{th}$. 
Through this learning strategy, the DRL algorithm reduces the probability of learning failure as well as achieves improved performance and fast convergence to the higher reward (please refer to Section~\ref{sec:exp}). 
\begin{algorithm}
\caption{Entropy-aware model initialization.}\label{alg1}
\SetAlgoLined
 \hspace*{1pt}\textbf{Input:}
 Task ($E$), Number of actors ($M$), Entropy threshold ($h_{th}$), Initialized model ($\pi_{0}$), Horizon ($T$)\\
 \hspace*{1pt}\textbf{Output:} 
 Model ($\pi_{init}$) for a DRL algorithm
 
 $i=0$
 
 \While{True}{
    \For{$j=1,2,\cdots,M$}{{Rollout with $\pi_{i}$ for $T$ in $E$\; 
    Store action selection probabilities;}}
    Compute mean entropy, $\hat{h}_{\pi_{i}}$, in Equation~\ref{eqn1}\;
    
  \eIf{$\hat{h}_{\pi_{i}}$ $>$ $h_{th}$}{
   $\pi_{init}\leftarrow\pi_{i}$\;
   break\;
   }{Set random seeds on current timestamp\;
  Reinitialize model, $\pi_{i+1}$\;
  $i\leftarrow i+1$\;}
 }
\end{algorithm}

\section{Experimental Results}\label{sec:exp}
In this section, we validated the effectiveness of the devised learning strategy. 
For the validation, we used the same experimental settings and tasks described in Section~\ref{sec:effect}. 
In this experiment, we set the entropy threshold ($h_{th}$) to 0.5.

\begin{figure}[!t]
  \centering
 \includegraphics[width=0.7\linewidth]{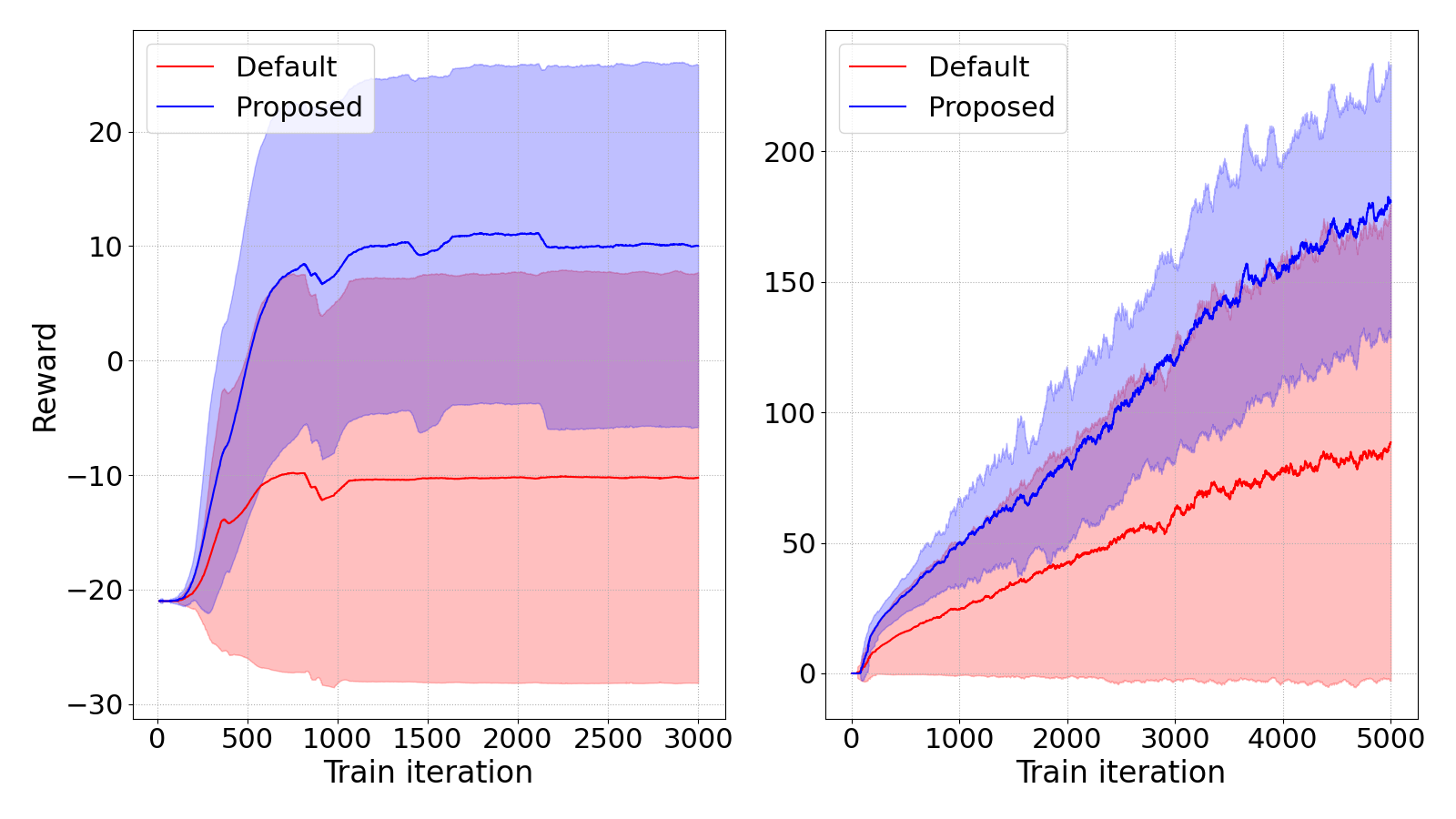}
 \caption{Comparison of the entropy-aware model initialization-based DRL with the conventional DRL for two tasks, (left) Pong and (right) Breakout.}
\label{fig_res1}
\end{figure}

In order to validate the effect of the entropy-aware model initialization, we considered 30 models initialized by the different random seeds for each task. 
Fig.~\ref{fig_res1} shows the rewards according to the training iterations for the Pong and Breakout tasks. 
In this figure, the red line represents the result for the conventional DRL (without the entropy-aware model initialization) denoted as ``Default'', and the blue line denotes the result for the proposed entropy-aware model initialization denoted as ``Proposed''.
We observe that the DRL with the proposed learning strategy outperforms the conventional DRL for both tasks in four aspects.
1) It restrains the learning failures, \textit{e.g.}, the learning failures for the ``Proposed'' are 6 and 1, but for the ``Default'' are 22, and 15, for Pong and Breakout, respectively; 2) It enhances the performance by 1.98 for Pong, and 2.04 times for Breakout, respectively; 3) It reduces the performance variations; 4) It enhances the learning speed.
Table~\ref{tab_comp} shows the overheads of the entropy-aware model initialization. The overheads are the number and time for the repetitive initialization. 
We can observe that the time overheads of the proposed strategy are less than a minute and do not differ much depending on the task. In contrast, the total training times are more than an hour, and the more complex the task, the longer the training time. Therefore, the overheads due to the proposed strategy become negligible as the task complexity grows.
\begin{table*}[!t]
\begin{center}
\caption{The average number and time of initialization by the entropy-aware model initialization and the total training time with the standard deviation.}
\vspace{0.3cm}
\label{tab_comp}
\begin{tabular}{|l||c|c|}
\hline
&
Pong &
Breakout\\ \Xhline{3\arrayrulewidth}
Average Number of Initialization (\#) & {\begin{tabular}[c]{@{}c@{}} 4.8 \\ (3.55) \end{tabular}} & {\begin{tabular}[c]{@{}c@{}} 5.0 \\ (4.56) \end{tabular}}\\ \hline
Average Time for Initialization (sec)  & {\begin{tabular}[c]{@{}c@{}} 52.465 \\ (33.74) \end{tabular}} &  {\begin{tabular}[c]{@{}c@{}} 55.224 \\ (46.246) \end{tabular}}\\\hline
Average Training Time (sec) & {\begin{tabular}[c]{@{}c@{}} 6,166.844 \\ (885.026) \end{tabular}} & {\begin{tabular}[c]{@{}c@{}} 10,278.073 \\ (1,475.044) \end{tabular}} \\\hline
\end{tabular}
\end{center}
\end{table*}

\section{Conclusion}\label{sec:con}
In this paper, we have conducted an experimental study to investigate the effect of initial entropy in the DRL framework focusing on the tasks with discrete action spaces. 
The critical observation is that the models with low initial entropy lead to frequent learning failures even with the easy tasks. 
These initial entropy values are biased towards low values. 
Moreover, we have observed that the initial entropy significantly varies depending on the task and the initial model weight through the experiments under the various tasks.
Inspired by the experimental observations, we have devised a learning strategy, so-called Entropy-aware model initialization, which repeatedly initializes the model and measures the entropy until the initial entropy exceeds a certain threshold. Its purpose is to provide a well-initialized model with a reinforcement learning algorithm.
Through the devised learning strategy, learning failure, performance, performance variation, and learning speed are all improved. Furthermore, it is practical since it is easy to implement and easy to apply with various reinforcement learning algorithms without modifying them.

\section*{Acknowledgement}
\noindent This work was partly supported by Electronics and Telecommunications Research Institute (ETRI) and Institute of Information \& Communications Technology Planning \& Evaluation (IITP) grant funded by the Korea government (MSIT) (21ZR1100, A Study of Hyper-Connected Thinking Internet Technology by autonomous connecting, controlling and evolving ways, No.2020-0-00004, Development of Previsional Intelligence based on Long-term Visual Memory Network).

\end{document}